\documentclass[sigconf]{acmart}
\usepackage{multirow}
\usepackage{enumitem}
\usepackage{balance}

\usepackage{color}
\definecolor{mygreen}{RGB}{112,173,71} 
\definecolor{myblue}{RGB}{0,112,192} 
\definecolor{myorange}{RGB}{255,141,65}

\AtBeginDocument{%
  \providecommand\BibTeX{{%
    \normalfont B\kern-0.5em{\scshape i\kern-0.25em b}\kern-0.8em\TeX}}}

\setcopyright{acmcopyright}
\copyrightyear{2022}
\acmYear{2022}
\setcopyright{acmcopyright}\acmConference[MM '22]{Proceedings of the 30th ACM
International Conference on Multimedia}{October 10--14, 2022}{Lisboa, Portugal}
\acmBooktitle{Proceedings of the 30th ACM International Conference on Multimedia
(MM '22), October 10--14, 2022, Lisboa, Portugal}
\acmPrice{15.00}
\acmDOI{10.1145/3503161.3547910}
\acmISBN{978-1-4503-9203-7/22/10}

\begin{document}

\title{X-CLIP: End-to-End Multi-grained Contrastive Learning for Video-Text Retrieval}


% \affiliation{%

  \author{
     Yiwei Ma$^{1*}$,\quad Guohai Xu$^{3}$,\quad Xiaoshuai Sun$^{12\dagger}$,\quad Ming Yan$^{3}$,\quad Ji Zhang$^{3}$,\quad Rongrong Ji$^{12}$ 
    }
     \affiliation{\institution{$^{1}$Media Analytics and Computing Lab, Department of Artificial Intelligence, School of Informatics,\\ Xiamen University, 361005 \country{China},  $^{2}$Institute of Artificial Intelligence, \\Xiamen University,  $^{3}$DAMO Academy, Alibaba Group.}}
%, $^{4}$Fujian Engineering Research Center of Trusted Artificial\\ Intelligence Analysis and Application, Xiamen University, 361005, China

    % \email{yiweima@stu.xmu.edu.cn, guohai.xgh@alibaba-inc.com, xssun@xmu.edu.cn,
    %  }
    % \email{ym119608@alibaba-inc.com,zj12214@alibaba-inc.com, rrji@xmu.edu.cn
    %  }
     
\def\authors{Yiwei Ma, Guohai Xu, Xiaoshuai Sun, Ming Yan, Ji Zhang, Rongrong Ji}

\renewcommand{\shortauthors}{Yiwei Ma et al.}

\begin{abstract}
  Video-text retrieval has been a crucial and fundamental task in multi-modal research. The development of video-text retrieval has been considerably promoted by large-scale multi-modal contrastive pre-training, which primarily focuses on coarse-grained or fine-grained contrast. However, cross-grained contrast, which is the contrast between coarse-grained representations and fine-grained representations, has rarely been explored in prior research. Compared with fine-grained or coarse-grained contrasts, cross-grained contrast calculate the correlation between coarse-grained features and each fine-grained feature, and is able to filter out the unnecessary fine-grained features guided by the coarse-grained feature during similarity calculation, thus improving the accuracy of retrieval. To this end, this paper presents a novel multi-grained contrastive model, namely X-CLIP, for video-text retrieval. However, another challenge lies in the similarity aggregation problem, which aims to aggregate fine-grained and cross-grained similarity matrices to instance-level similarity. To address this challenge, we propose the Attention Over Similarity Matrix (AOSM) module to make the model focus on the contrast between essential frames and words, thus lowering the impact of unnecessary frames and words on retrieval results. With multi-grained contrast and the proposed AOSM module, X-CLIP achieves outstanding performance on five widely-used video-text retrieval datasets, including MSR-VTT (\emph{49.3 R@1}), MSVD (\emph{50.4 R@1}), LSMDC  (\emph{26.1 R@1}), DiDeMo  (\emph{47.8 R@1}) and ActivityNet (\emph{46.2 R@1}). It outperforms the previous state-of-the-art by \emph{+6.3\%}, \emph{+6.6\%}, \emph{+11.1\%}, \emph{+6.7\%}, \emph{+3.8\%} relative improvements on these benchmarks, demonstrating the superiority of multi-grained contrast and AOSM. Code is available at \url{https://github.com/xuguohai/X-CLIP}.
\end{abstract}



\maketitle

\section{Introduction} \label{sec:intro}

Video-text retrieval (VTR) is a multi-modal task, which aims to find the most relevant video/text based on the text/video query. With the explosive growth of videos on the Internet, VTR has attracted increasing interests and served as an important role in people's daily life. Recent years have witnessed the rapid development of VTR, which is supported by a series of pre-training multi-modal models \cite{radford2021learning,lei2021less,bain2021frozen}, innovative retrieval methods \cite{yu2018joint,zhu2020actbert,lei2021less,liu2019use,gabeur2020multi,dzabraev2021mdmmt,mithun2018learning,zhang2018cross,liu2021hit,dong2019dual,bertasius2021space,arnab2021vivit,luo2021clip4clip,DBLP:conf/sigir/JinZZZHZ21,DBLP:conf/sigir/YangD0W0C20,DBLP:conf/cvpr/WangZ021} and video-text benchmarks \cite{caba2015activitynet,xu2016msr,chen2011collecting,rohrbach2015long,anne2017localizing}.

Recently, with great success in large-scale contrastive language-image pre-training, VTR has also achieved great progress.  Specifically, with 400M image-text pairs for training, CLIP \cite{radford2021learning} can embed the images and sentences into the shared semantic space for similarity calculation. Furthermore, CLIP4Clip \cite{luo2021clip4clip} transfers the image-text knowledge of CLIP to the VTR task, resulting in significant performance improvements on several video-text retrieval datasets. However, CLIP and CLIP4Clip embed the whole sentence and image/video into textual and visual representations, thus lacking the ability to capture fine-grained interactions. To this end, some previous works \cite{yao2021filip,lee2018stacked} propose fine-grained contrastive frameworks, which consider the contrast between each word of the sentence and each frame of the video. Moreover, TACo \cite{yang2021taco} introduces token-level and sentence-level loss to consider both fine-grained and coarse-grained contrast. Although they have shown promising advances on the VTR task, cross-modality semantic contrast still needs to be systematically explored.

\begin{figure}
\centering 
  \includegraphics[width=0.9\columnwidth]{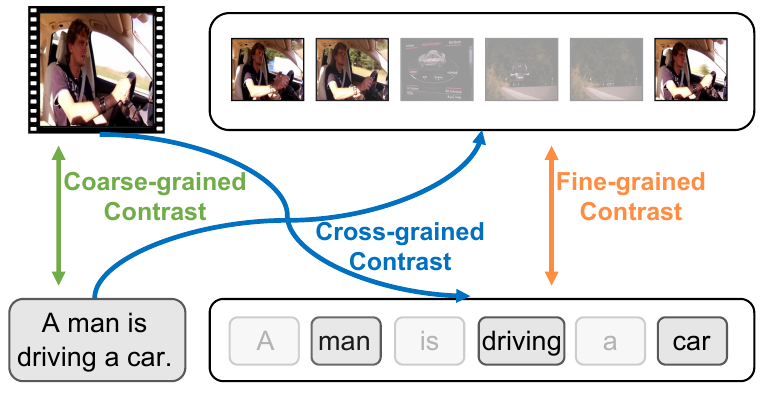}
  \vspace{-0.3cm}
  \caption{X-CLIP aims for improving video-text retrieval performance via multi-grained contrastive learning, including \textcolor{myorange}{fine-grained (frame-word)}, \textcolor{mygreen}{coarse-grained (video-sentence)}  and \textcolor{myblue}{cross-grained (video-word, sentence-frame)} contrast. The transparency of words and frames represents the degree of relevance to query.}
  \vspace{-0.3cm}
  \label{fig:intro}
\end{figure}

As shown in Fig. \ref{fig:intro}, a video is composed of multiple frames, and a sentence consists of several words. Video and sentence are usually redundant, which may contain some unnecessary frames or unimportant words. Concretely, given a specific video or sentence query, unnecessary frames or unimportant words refer to the candidates with low relevance to the query (\emph{i.e.,} light-colored frames and words in Fig. \ref{fig:intro}). However, most current works mainly focus on coarse-grained contrast \cite{radford2021learning,luo2021clip4clip}, fine-grained contrast \cite{yao2021filip,lee2018stacked} or both \cite{yang2021taco}, which are inefficient in filtering out these unnecessary frames and words. Specifically, coarse-grained contrast calculates the similarity between video-level and sentence-level features, and fine-grained contrast calculates the similarity between frame-level and word-level features. To this end, we ask: \emph{How to effectively filter out unnecessary information during retrieval?} To answer this question, we propose the cross-grained contrast, which calculates the similarity score between the coarse-grained features and each fine-grained feature. As shown in Fig. \ref{fig:intro}, with the help of the coarse-grained feature, unimportant fine-grained features will be filtered out and important fine-grained features will be up-weighted. However, challenges in cross-grained contrast arise from aggregating similarity matrices to instance-level similarity scores. A naive and easy method is to use \emph{Mean-Max} strategy \cite{yao2021filip,khattab2020colbert,santhanam2021colbertv2,khattab2021relevance} to calculate the instance-level similarity score after obtaining the similarity matrix. However, the conventional \emph{Mean-Max} strategy is not conducive to filtering out the unnecessary information in videos and sentences during retrieval. On one hand, \emph{Mean} applies the same weight to all frames and words, so the contrast between unnecessary frames and unimportant words may harm the retrieval performance. On the other hand, \emph{Max} only considers the most important frame and word, ignoring other critical frames and words.

Based on the above analysis, in this paper, we propose an end-to-end multi-grained contrast model, namely \emph{\textbf{X-CLIP}}, for video-text retrieval. Specifically, X-CLIP first adopts modality-specific encoders to generate multi-grained visual and textual representations and then considers multi-grained contrast of features (\emph{i.e.,} video-sentence, video-word, sentence-frame, and frame-word) to obtain multi-grained similarity scores, vectors, and matrices. To effectively filter out the unnecessary information and obtain meaningful instance-level similarity scores, the AOSM module of X-CLIP conducts the attention mechanism over the similarity vectors/matrices. Different from the conventional \emph{Mean-Max} strategy, our proposed AOSM module dynamically considers the importance of each frame in the video and each word in the sentence, so the adverse effects of unimportant words and unnecessary frames on retrieval performance are reduced.

To validate the effectiveness of our proposed X-CLIP, we conduct extensive experiments on five widely-used video-text retrieval benchmarks and achieve signiﬁcantly better performance than previous approaches. Specifically, our X-CLIP achieves 49.3 R@1 on MSR-VTT (\emph{i.e.,} 6.3\% relative improvement, 2.9\% absolute improvement over the previous state-of-the-art approach). Besides, our proposed X-CLIP achieves 50.4 R@1, 26.1 R@1, 47.8 R@1, 46.2 R@1 on the MSVD, LSMDC, DiDeMo and ActivityNet datasets, respectively, which outperforms the previous SOTA method by +6.6\% (+3.1\%), +11.1\% (+2.6\%), +6.7\% (+3.0\%), +3.8\% (+1.7\%) on relative (absolute) improvement.

\section{Related Works}

\subsection{Vision-Language Pre-Training}
With the success of self-supervised pre-training such as BERT~\citep{devlin2018bert} in NLP, vision-language pre-training on large-scale unlabeled cross-modal data has attracted growing attention~\citep{lu2019vilbert,xu-etal-2021-e2e,tan2019lxmert,li2020oscar,yu2020ernie,li2021align,radford2021learning,jia2021scaling,sun2019videobert,li2020hero}. One line of work such as LXMERT~\citep{tan2019lxmert}, OSCAR~\citep{li2020oscar} and ALBEF~\citep{li2021align} focuses on pre-training on enormous image-text pairs data, and obtains significant improvement in a variety of vision-and-language tasks. To better cope with the image-text retrieval tasks, contrastive language-image pre-training methods such as CLIP~\citep{radford2021learning}, ALIGN~\citep{jia2021scaling} and WenLan~\citep{huo2021wenlan} have been proposed, by leveraging billion-scale image-text pairs data from the web with a dual-stream Transformer. Due to the great advantage of CLIP for visual representation learning, some recent work such as CLIP4Clip~\citep{luo2021clip4clip} has also begun to transfer the knowledge of CLIP to video-text retrieval tasks and obtained new state-of-the-art results. The other line of work such as VideoBERT~\citep{sun2019videobert}, HERO~\citep{li2020hero} and Frozen in Time~\citep{bain2021frozen} directly collects video-text pairs data for video-language pre-training, by further considering the temporal information in videos. However, the scale of the video-language pre-training dataset is much smaller than image-text pre-training since the process of video-text dataset collection is much more expensive. In this work, we follow the line of CLIP4Clip~\citep{luo2021clip4clip}, which enhances video-text retrieval by borrowing the ability of visual representation learning from contrastive image-text pre-training. Different from CLIP4Clip~\citep{luo2021clip4clip}, we design a multi-grained video-text alignment function to better align the video-text semantics. 

\subsection{Video-Text Retrieval}
Video-text retrieval is a popular but challenging task, which involves cross-modal fusion of multiple modalities and additional understanding of temporal information in videos. Traditional video-text retrieval methods tend to design task-specific or modality-specific fusion strategies for cross-modal learning from offline extracted video and text features~\citep{yu2017end,gabeur2020multi,he2021improving,liu2021hit,patrick2020support,jang2017tgif,le2020hierarchical}, including face recognition/object recognition/audio processing. However, they are limited by the pre-extracted single modal features, since these features are not properly learnt for the target downstream tasks. Recently, the paradigm of end-to-end video-text retrieval by training models directly from raw video/text has gained large popularity. For example, MIL-NCE~\citep{miech2020end} adopts Multiple Instance Learning and Noise Contrastive Estimation for end-to-end video representation learning, which addresses visually misaligned narrations from uncurated videos. ClipBERT~\citep{lei2021less} proposes to sparsely sample video clips for end-to-end training to obtain clip-level predictions, while Frozen in Time~\citep{bain2021frozen} uniformly samples video frames and conducts end-to-end training on both image-text and video-text pairs data. CLIP4Clip~\citep{luo2021clip4clip} transfers the knowledge of CLIP to end-to-end video-text retrieval and investigates three similarity calculation approaches for video-sentence contrastive learning. However, cross-grained (\emph{i.e.,} video-word and sentence-frame) contrast is also critical, which has rarely been explored in previous works. We propose the first work of multi-grained contrastive learning for end-to-end video-text retrieval, by considering all the video-sentence, video-word, sentence-frame, and frame-word contrasts.

\subsection{Multi-Grained Contrastive Learning}

Recently, contrastive learning \cite{chen2020simple,he2020momentum,chen2020improved,chen2021empirical} has been a popular topic in deep learning community. CLIP \cite{radford2021learning} implements the idea of contrastive learning based on a large number of image-text pairs, achieving outstanding performance on several multi-modal downstream tasks~\citep{zhang2021rstnet,ji2022knowing,ma2022knowing,zhu2022seqtr,ji2021improving,he2022pixelfolder}. To achieve fine-grained contrastive learning, FILIP \cite{yao2021filip} contrasts the patch in the image with the word in the sentence, achieving fine-grained semantic alignment. TACo \cite{yang2021taco} proposes token-level and sentence-level losses to include both fine-grained and coarse-grained contrasts. Although contrastive learning has been widely used in multi-modal pre-training, cross-grained contrast has rarely been explored in previous works, which is also critical for semantic alignment. Therefore, we propose a multi-grained contrastive learning method for video-text retrieval, which aims to achieve multi-grained semantic alignment.

\section{METHODOLOGY}

\begin{figure*}
\centering 
  \includegraphics[width=1.5\columnwidth]{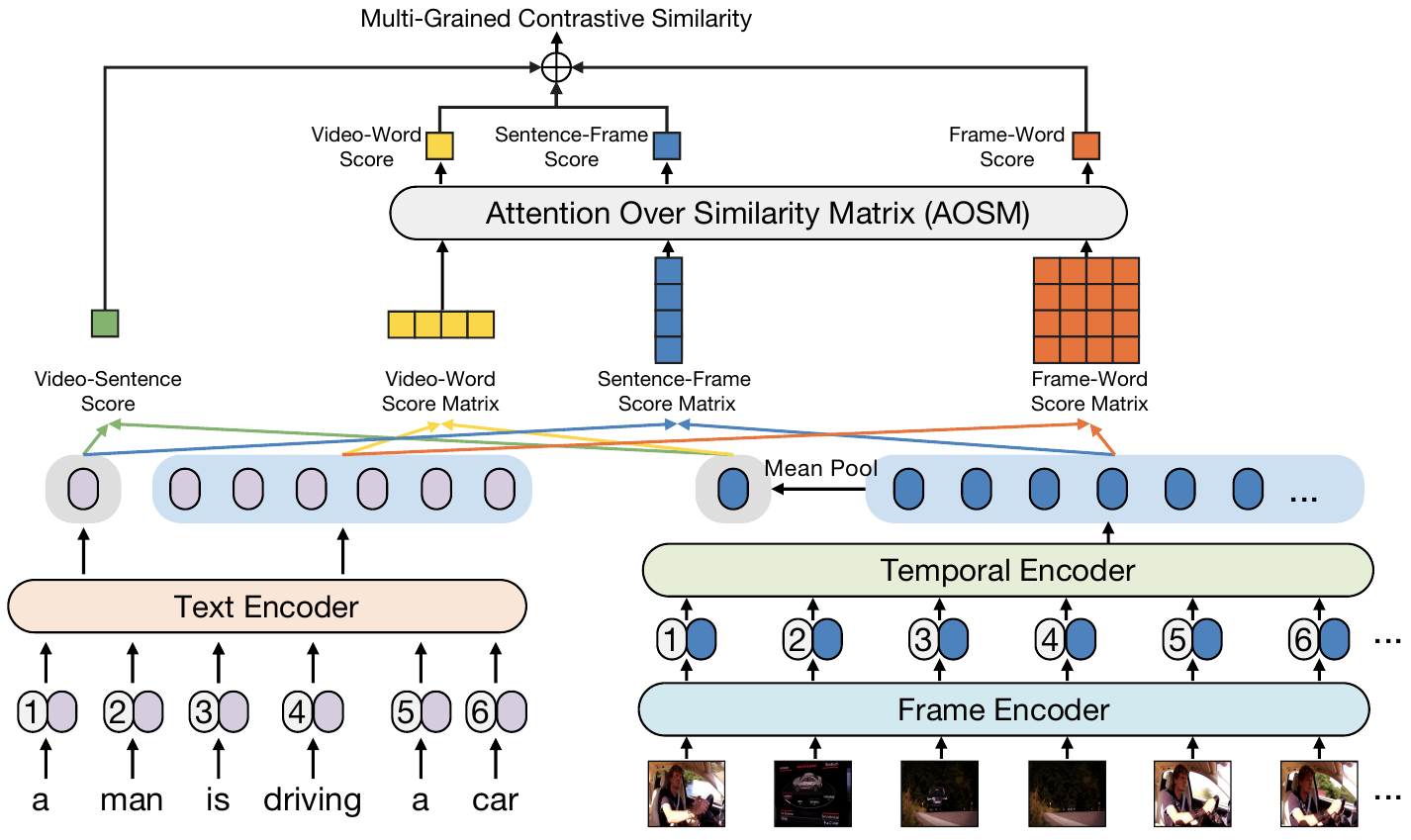}
  \vspace{-0.3cm}
  \caption{Illustration of the proposed \emph{X-CLIP} model. The input sentences are processed by the text encoder to generate coarse-grained and fine-grained textual representations. The input video is sampled into ordinal frames and these frames are fed into the frame encoder to generate frame-level representations. The frame-level representations are then fed into the temporal encoder to capture the temporal relationships. The outputs of the temporal encoder are fine-grained visual representations, and the coarse-grained visual representation is obtained by averaging all these fine-grained features. Based on these representations, we calculate the video-sentence, video-word, sentence-frame, and frame-word similarity score.}
  \vspace{-0.2cm}
  \label{fig:overview}
\end{figure*}

In this section, we elaborate each component of our proposed X-CLIP, whose architecture is shown in Fig. \ref{fig:overview}. Specifically, we first introduce how to extract the multi-grained visual and textual representations in Sec. \ref{sec:feature}. We then explain the multi-grained contrastive learning based on these feature representations in Sec. \ref{sec:multigraied}, which aims to obtain multi-grained contrast scores, vectors, and matrices. We also introduce how to aggregate the similarity vectors/matrices to the instance-level similarity score in Sec. \ref{sec:attention}. Finally, we describe the similarity calculation and objective function for video-text retrieval in Sec. \ref{sec:similarity} and \ref{sec:objective}, respectively.

\subsection{Feature Representation} \label{sec:feature}

\subsubsection{Frame-level Representation}
For a video $\hat{v}_i \in \mathbf{\hat{V}}$, we first sample video frames using the sampling rate of 1 frame per second (FPS). Frame encoder is used to process these frames to obtain frame-level features, which is a standard vision transformer (ViT) with 12 layers. Following the previous work \cite{luo2021clip4clip}, we initialize our frame encoder with the public CLIP \cite{radford2021learning} checkpoints. The architecture of ViT is the same as the transformer \cite{vaswani2017attention} encoder in natural language processing (NLP), except ViT introduces a visual tokenization process to convert video frames into discrete token sequences. The discrete token sequence, which is prepended with a [CLS] token, is then fed into the Transformer of ViT. The [CLS] tokens from the last layer are extracted as the frame-level features $\bar{v}_{(i,j)} \in \mathbf{\bar{V}}_i$. 

\subsubsection{Visual Representation}
However, $\bar{v}_{(i,j)} \in \mathbf{\bar{V}}_i$ are extracted from separate frames, without considering the interaction among frames. Therefore, we further propose a temporal encoder with temporal position embedding $\mathbf{P}$, which is a set of predefined parameters, to model the temporal relationship. To be specific, the temporal encoder is also a standard transformer with 3 layers, which can be formulated as:
\begin{equation}
    \mathbf{V}_i=TransEnc\left(\mathbf{\bar{V}}_i + \mathbf{P}\right),
\label{eq:trans}
\end{equation}
where $\mathbf{V}_i=[v_{(i,1)},v_{(i,2)},v_{(i,3)},...,v_{(i,n)}]$ is the final frame-level (fine-grained) visual features for the video $\hat{v}_i$, $n$ is the number of frames in the video $\hat{v}_i$. To obtain video-level (coarse-grained) visual feature $v^{\prime}_i \in \mathbb{R}^{dim}$, all frame-level features of the video  ${v}_i $ are averaged, which can be formulate as:
\begin{equation}
    v^{\prime}_i=\frac{1}{n}\sum_j^n v_{(i,j)}.
\label{eq:trans}
\end{equation}

\subsubsection{Textual Representation}
Given a sentence, we directly use the text encoder of CLIP to generate the textual representation, which is also initialized by the public checkpoints of CLIP \cite{radford2021learning}. Specifically, it is a transformer encoder, which consists of multi-head self-attention and feed-forward networks. The transformer consists of 12 layers and 8 attention heads. The dimension of the query, key, and value features is 512. The tokenizer used in the experiment is lower-cased byte pair encoding (BPE) \cite{sennrich2015neural} with a 49,152 vocab size. Before being fed into the text encoder, the textual token sequence is padded with [BOS] and [EOS] at the beginning and end, respectively. The sentence-level (coarse-grained) textual feature $t^{\prime}_i \in \mathbb{R}^{dim}$ and word-level (fine-grained) textual features $\mathbf{T}_i=[t_{(i,1)},t_{(i,2)},t_{(i,3)},...,t_{(i,m)}]$ are the outputs of the [EOS] token and corresponding word tokens from the final layer of text encoder, where $m$ is the length of the sentence.

\subsection{Multi-Grained Contrastive Learning} \label{sec:multigraied}

Previous VTR works \cite{luo2021clip4clip,lee2018stacked} focus on fine-grained and coarse-grained contrastive learning, which include video-sentence and frame-word contrasts. However, as explained in Sec. \ref{sec:intro}, cross-grained (\emph{i.e.,} video-word and sentence-frame) contrast is explicit to filter out the unnecessary information in the video and sentence. Therefore, different from previous works \cite{luo2021clip4clip,lee2018stacked,yao2021filip}, which only focus single-grained contrast, X-CLIP is a multi-grained contrastive framework for VTR.

\subsubsection{Video-Sentence Contrast}

Given the video-level representation $v^{\prime} \in \mathbb{R}^{dim}$ and sentence-level representation $t^{\prime} \in \mathbb{R}^{dim}$, we use matrix multiplication to evaluate the similarity between video and sentence, which can be formulated as:
\begin{equation} 
    S_{V-S}=(v^{\prime})^\intercal (t^{\prime}),
\label{eq:video_sentence_simi}
\end{equation}
where $S_{V-S} \in \mathbb{R}^1$ is the video-sentence similarity score. \footnote{For clarity and simplicity, we have omitted the frame (word) index and video (sentence) index of visual (textual) representations.}

\subsubsection{Video-Word Contrast} \label{sec:video_word}

For the given video-level representation $v^{\prime} \in \mathbb{R}^{dim}$ and word-level representation vector $\mathbf{T}  \in \mathbb{R}^{m \times dim}$, we use matrix multiplication to calculate the similarity between the video representation and each word representation, which can be represented as follows:
\begin{equation}
    S_{V-W}=(\mathbf{T} v^{\prime})^\intercal,
\label{eq:video_word_simi}
\end{equation}
where $S_{V-W} \in \mathbb{R}^{1 \times m}$ is the similarity vector between video and each word in the sentence, $m$ is the length of the sentence.

\subsubsection{Sentence-Frame Contrast} \label{sec:frame_sentence}

Similar to Video-Word Contrast, we can calculate the similarity between the sentence representation $t^{\prime} \in \mathbb{R}^{dim}$ and each frame representation $\bar{\mathbf{V}}  \in \mathbb{R}^{n \times dim}$ based on matrix multiplication, which can be formulated as follows:
\begin{equation}
    S_{F-S}=\bar{\mathbf{V}}   t^{\prime},
\label{eq:frame_sentence_simi}
\end{equation}
where $S_{F-S} \in \mathbb{R}^{n \times 1}$ is the similarity vector between the sentence and each frame of a video, $n$ is the number of frames in the video.

\subsubsection{Frame-Word Contrast} \label{sec:frame_word}

The fine-grained similarity matrix between word representations and frame representations can be also obtained using the matrix multiplication:
\begin{equation}
    S_{F-W}=\bar{\mathbf{V}}   \mathbf{T}^\intercal,
\label{eq:frame_word_simi}
\end{equation}
where $S_{F-W} \in \mathbb{R}^{n \times m}$ is the fine-grained similarity matrix, $n$ and $m$ are the number of frames and words, respectively.

\subsection{Attention Over Similarity Matrix (AOSM)} \label{sec:attention}

To obtain the instance-level similarity, we fuse the similarity vector/matrix in Eq. \ref{eq:video_word_simi}, Eq. \ref{eq:frame_sentence_simi} and Eq. \ref{eq:frame_word_simi}. 
As discussed in Sec. \ref{sec:intro}, \emph{Mean-Max} strategies \cite{yao2021filip,khattab2020colbert,santhanam2021colbertv2,khattab2021relevance} ignore the importance of different frames and words. To address this issue, we propose the \emph{Attention Over Similarity Matrix (AOSM)} module, where scores in similarity vectors/matrices will be given different weights during aggregation.

Specifically, given the similarity vectors $S_{V-W} \in \mathbb{R}^{1 \times m}$ and $S_{F-S} \in \mathbb{R}^{n \times 1}$, we first use \emph{Softmax} to obtain the weights for the similarity vector, where scores for the fine-grained features related to the query will be given high weights. Then, we aggregate these similarity scores based on the obtained weights, which can be formulated as follows:
\begin{equation}
    S^{\prime}_{V-W}=\sum _{i=1} ^m \frac{exp(S_{V-W (1,i)}/\tau) }{\sum _{j=1} ^m exp(S_{V-W (1,j)}/\tau)} S_{V-W (1,i)}, 
\label{eq:attention_list1}
\end{equation}
\begin{equation}
    S^{\prime}_{F-S}=\sum _{i=1} ^n \frac{exp(S_{F-S (i,1)}/\tau) }{\sum _{j=1} ^n exp(S_{F-S (j,1)}/\tau)} S_{F-S (i,1)},
\label{eq:attention_list2}
\end{equation}
where $\tau$ is the temperature parameter of Softmax.

Since the fine-grained similarity matrix $S_{F-W} \in \mathbb{R}^{n \times m}$ contains the similarity scores of $n$ frames and $m$ words, we perform attention operations on the matrix twice. The first attention aims to get fine-grained video-level and sentence-level similarity vectors, which can be formulated as follows:
\begin{equation}
    S_{vid}=\sum _{i=1} ^n \frac{exp(S_{F-W (i,*)}/\tau) }{\sum _{j=1} ^n exp(S_{F-W (j,*)}/\tau)} S_{F-W (i,*)},
\label{eq:attention_video}
\end{equation}
\begin{equation}
    S_{sen}=\sum _{i=1} ^m \frac{exp(S_{F-W (*,i)}/\tau) }{\sum _{j=1} ^m exp(S_{F-W (*,j)}/\tau)} S_{F-W (*,i)}, 
\label{eq:attention_sentence}
\end{equation}
where $*$ represents all content in the dimension, $S_{vid} \in \mathbb{R}^{1 \times m}$ and $S_{sen} \in \mathbb{R}^{n \times 1}$ are the video-level and sentence-level similarity vector, respectively. Specifically, $S_{vid} \in \mathbb{R}^{1 \times m}$ shows the similarity score between the video and $m$ words in the sentence.  $S_{sen} \in \mathbb{R}^{n \times 1}$ represents the similarity score between the sentence and $m$ frames in the video.

To obtain fine-grained instance-level similarity scores, we conduct the second attention operation on the video-level vector $S_{vid} \in \mathbb{R}^{1 \times m}$ and sentence-level similarity vector $S_{sen} \in \mathbb{R}^{n \times 1}$, which can be represented as follows: 
\begin{equation}
    S^{\prime}_{vid}=\sum _{i=1} ^m \frac{exp(S_{vid (1,i)}/\tau) }{\sum _{j=1} ^m exp(S_{vid (1,j)}/\tau)} S_{vid (1,i)}, 
\label{eq:attention_video2}
\end{equation}
\begin{equation}
    S^{\prime}_{sen}=\sum _{i=1} ^n \frac{exp(S_{sen (i,1)}/\tau) }{\sum _{j=1} ^n exp(S_{sen (j,1)}/\tau)} S_{sen (i,1)},
\label{eq:attention_sentence2}
\end{equation}
where $S^{\prime}_{vid} \in \mathbb{R}^1$ and $S^{\prime}_{sen} \in \mathbb{R}^1$ are the instance-level similarities. We use the average value as the fine-grained similarity score:
\begin{equation}
    S^{\prime}_{F-W}=(S^{\prime}_{vid}+S^{\prime}_{sen})/2.
\label{eq:final_simi}
\end{equation}

\begin{table*}[]
\caption{Retrieval performance comparison to SOTAs on the MSR-VTT dataset.}
\vspace{-0.3cm}
\begin{tabular}{l|ccccc|ccccc}
\hline
           & \multicolumn{5}{c|}{Text-to-Video Retrieval}                                  & \multicolumn{5}{c}{Video-to-Text Retrieval}                                  \\
           \hline
Model      & R@1↑          & R@5↑          & R@10↑         & MdR↓         & MnR↓          & R@1↑          & R@5↑          & R@10↑
& MdR↓         & MnR↓          \\ \hline
CE \cite{liu2019use}        & 20.9          & 48.8          & 62.4          & 6.0          & 28.2          & 20.6          & 50.3          & 64.0          & 5.3          & -             \\
MMT  \cite{gabeur2020multi}      & 26.6          & 57.1          & 69.6          & 4.0          & 24.0          & 27.0          & 57.5          & 69.7          & 3.7          & -             \\
AVLnet \cite{rouditchenko2020avlnet}    & 27.1          & 55.6          & 66.6          & 4.0          & -             & 28.5          & 54.6          & 65.2          & 4.0          & -             \\
SSB \cite{patrick2020support}       & 30.1          & 58.5          & 69.3          & 3.0          & -             & 28.5          & 58.6          & 71.6          & 3.0          & -             \\
MDMMT \cite{dzabraev2021mdmmt}     & 38.9          & 69.0          & 79.7          & \textbf{2.0}          & 16.5          & -             & -             & -             & -            & -             \\
Frozen \cite{bain2021frozen}    & 31.0          & 59.5          & 70.5          & 3.0          & -             & -             & -             & -             & -            & -             \\
HiT  \cite{liu2021hit}      & 30.7          & 60.9          & 73.2          & 2.6          & -             & 32.1          & 62.7          & 74.1          & 3.0          & -             \\
TT-CE+  \cite{croitoru2021teachtext}   & 29.6          & 61.6          & 74.2          & 3.0          & -             & 32.1          & 62.7          & 75.0          & 3.0          & -             \\
CLIP-straight  \cite{portillo2021straightforward}     & 31.2          & 53.7          & 64.2          & 4.0          & -             & 27.2          & 51.7          & 62.6          & 5.0          & -             \\
CLIP4Clip-MeanP (ViT-B/32) \cite{luo2021clip4clip}  & 43.1          & 70.4          & 80.8          & \textbf{2.0} & 16.2          & 43.1          & 70.5          & 81.2          & \textbf{2.0} & 12.4          \\ 
CLIP4Clip-seqLSTM (ViT-B/32) \cite{luo2021clip4clip}  & 42.5          & 70.8          & 80.7          & \textbf{2.0} & 16.7          & 42.8          & 71.0          & 80.4          & \textbf{2.0} & 12.3          \\ 
CLIP4Clip-seqTransf (ViT-B/32) \cite{luo2021clip4clip}  & 44.5          & 71.4          & 81.6          & \textbf{2.0} & 15.3          & 42.7          & 70.9          & 80.6          & \textbf{2.0} & 11.6          \\ 
CLIP4Clip-tightTransf (ViT-B/32) \cite{luo2021clip4clip}  & 40.2          & 71.5          & 80.5          & \textbf{2.0} & 13.4          & 40.6          & 69.5          & 79.5          & \textbf{2.0} & 13.6          \\ 
CLIP4Clip-MeanP (ViT-B/16) \cite{luo2021clip4clip}  & 45.3 & 73.3 & 83.0 & \textbf{2.0} & 13.0 & 44.8 & 73.2 & 82.2 & \textbf{2.0} & 9.6  \\    
CLIP4Clip-seqLSTM (ViT-B/16) \cite{luo2021clip4clip}  & 44.3 & 72.0 & 82.2 & \textbf{2.0} & 13.7 & 44.3 & 73.4 & 82.4 & \textbf{2.0} & 10.3  \\   
CLIP4Clip-seqTransf (ViT-B/16) \cite{luo2021clip4clip}  & 46.4 & 72.1 & 82.0 & \textbf{2.0} & 14.7 & 45.4 & 73.4 & 82.4 & \textbf{2.0} & 10.7  \\  
CLIP4Clip-tightTransf (ViT-B/16) \cite{luo2021clip4clip}  & 42.9 & 71.7 & 81.5 & \textbf{2.0} & 13.3 & 41.9 & 71.0 & 80.7 & \textbf{2.0} & 10.1 
\\ \hline
X-CLIP (ViT-B/32)         & 46.1          & 73.0          & 83.1          & \textbf{2.0} & 13.2          & 46.8          & 73.3          & 84.0          & \textbf{2.0} & 9.1          \\
X-CLIP (ViT-B/16) & \textbf{49.3} & \textbf{75.8} & \textbf{84.8} & \textbf{2.0} & \textbf{12.2} & \textbf{48.9} & \textbf{76.8} & \textbf{84.5} & \textbf{2.0} & \textbf{8.1} \\ \hline
\end{tabular}
\label{tab:mstvtt_performance}
\end{table*}

\subsection{Similarity Calculation} \label{sec:similarity}

The similarity score $s(v_i,t_j)$ measures the semantic similarity between the two instances. Different from the previous work \cite{luo2021clip4clip} that only consider the coarse-grained contrast, our proposed X-CLIP adopt multi-grained contrast during retrieval. Therefore, the final similarity score $s(v_i,t_j)$ of X-CLIP contains multi-grained contrastive similarity scores, which can be represented as follows:
\begin{equation}
    s(v_i,t_j)=(S_{V-S}+S^{\prime}_{V-W}+S^{\prime}_{F-S}+S^{\prime}_{F-W})/4.
\label{eq:simi}
\end{equation}

\subsection{Objective Function} \label{sec:objective}
During training, given a batch of $B$ video-text pairs, the model will generate a $B \times B$ similarity matrix. We adopt the symmetric InfoNCE loss over the similarity matrix to optimize the retrieval model, which can be formulated as:
\begin{equation}
    \mathcal{L}_{v 2 t}=-\frac{1}{B} \sum_{i=1}^{B} \log \frac{\exp \big(s(v_{i}, t_{i})\big)}{\sum_{j=1}^{B} \exp \big(s(v_{i}, t_{j})\big)},
\label{eq:t2v}
\end{equation}
\begin{equation}
    \mathcal{L}_{t 2 v}=-\frac{1}{B} \sum_{i=1}^{B} \log \frac{\exp \big(s(v_{i}, t_{i})\big)}{\sum_{j=1}^{B} \exp \big(s(v_{j}, t_{i})\big)},
\label{eq:v2t}
\end{equation}
\begin{equation}
    \mathcal{L}=\mathcal{L}_{v 2 t}+\mathcal{L}_{t 2 v}.
\label{eq:loss}
\end{equation}

\section{EXPERIMENTS}

\subsection{Datasets}

\noindent\textbf{MSR-VTT} \cite{xu2016msr} is a popular video-text retrieval dataset, which contains 10,000 videos and 200,000 captions. The length of videos in this dataset ranges from 10 to 32 seconds. In this paper, we adopt the widely-used `Training-9K' split, where 9,000 videos and 180,000 captions are used for training and the rest are used for testing.

\noindent\textbf{MSVD}
 \cite{chen2011collecting} contains 1,970 videos, the duration of which vary from 1 to 62 seconds. Each video is annotated with 40 English captions. We use 1,200, 100, 670 videos for training, validating, and testing. 

\noindent\textbf{LSMDC} \cite{rohrbach2015long} is a dataset that contains 118,081 videos and captions. The duration of each video ranges from 2 to 30 seconds. We adopt 109,673, 7,408, and 1,000 videos for training, validating, and testing.

\noindent\textbf{DiDeMo}  \cite{anne2017localizing} contains 10,000 videos and 40,000 captions. Following previous works \cite{liu2019use,lei2021less,bain2021frozen}, all captions of a video are concatenated together during video-paragraph retrieval.

\noindent\textbf{ActivityNet} \cite{caba2015activitynet} contains 20,000 YouTube videos, which are annotated temporally. Following previous works \cite{luo2021clip4clip,sun2019learning,gabeur2020multi}, all captions of a video are also concatenated together during video-paragraph retrieval for fair comparison.

\subsection{Experimental Settings}

\subsubsection{Implementation Details}
We conduct the experiments on 4 NVIDIA Tesla V100 32GB GPUs using the PyTorch library. Following the previous work \cite{luo2021clip4clip}, the text encoder and frame encoder of X-CLIP are initialized by the public CLIP checkpoints.  We use the Adam optimizer \cite{kingma2015adam} to optimize the X-CLIP and decay the learning rate using a cosine schedule strategy \cite{loshchilov2016sgdr}. Since the parameters of the text encoder and frame encoder are initialized from the public CLIP checkpoints, we adopt different learning rates for different modules. Specifically, the initial learning rate for text encoder and frame encoder is 1e-7, and the initial learning rate for other modules is 1e-4. We set the max token length, max frame length, batch size, and the training epoch to 32, 12, 300, and 3 for MSR-VTT, MSVD, and LSMDC datasets. Since videos and captions in DiDeMo and ActivityNet are longer and more complex, we set the max token length, max frame length, and the training epoch to 64, 64, and 20. Due to the limitation of GPU memory, we also reduce the batch size of DiDeMo and ActivityNet to 64. We conduct ablation, quantitative and qualitative experiments on the MSR-VTT dataset, it is more popular and competitive compared with other datasets. The base model of X-CLIP is ViT-B/32 if not specified. In order to enhance the expression ability of the model, we adopt linear embedding during calculating the video-sentence and frame-word similarity scores, which are initialized with the identity matrices. Besides, we also use the FC layers which are initialized with the identity matrices on similarity scores to enhance the modeling ability of the model.

\subsubsection{Evaluation Protocols}

To evaluate the retrieval performance of our proposed model, we use recall at Rank K (R@K, higher is better), median rank (MdR, lower is better), and mean rank (MnR, lower is better) as retrieval metrics, which are widely used in previous retrieval works \cite{yu2018joint,zhu2020actbert,lei2021less,liu2019use,gabeur2020multi,dzabraev2021mdmmt,mithun2018learning,zhang2018cross,liu2021hit,dong2019dual,bertasius2021space,arnab2021vivit,luo2021clip4clip}.

\subsection{Performance Comparison}
We compare X-CLIP against the previous works on MSR-VTT, MSVD, LSMDC, DiDeMo, and ActivityNet. X-CLIP achieves the SOTA results on all five datasets with significant improvements.

For the MSR-VTT dataset, the performance comparison is shown in Tab. \ref{tab:mstvtt_performance}. By analyzing the table, we gain the following observations:

\begin{itemize}[leftmargin=*]
    \item Benefiting from the large-scale image-text pre-training, both CLIP4Clip and our model X-CLIP can obtain significant gains in performance compared with all the baselines. The consistent improvements verify that it is important to adopt end-to-end finetuning to realize the full potential of the image-text pre-trained model on video-text retrieval.
    
    \item Compared with the strongest competitor (\emph{i.e.,} CLIP4Clip-seqTransf), X-CLIP obtains 49.3 R@1 (6.3\% relative improvement, 2.9\% absolute improvement) in the text-to-video retrieval task and 48.9 R@1 (7.7\% relative improvement, 3.5\% absolute improvement) in the video-to-text retrieval task by employing CLIP(ViT-B/16) as pre-trained model. This can be attributed to that our proposed cross-grained contrast and the AOSM module are critical to reducing the bad effects of unnecessary frames and unimportant words.

    \item Compared to all the other state-of-the-arts, our model with ViT-B/16 achieves the best performance in all metrics. Surprisingly, our model with the ViT-B/32 can even achieve comparable performance to CLIP4Clip with ViT-B/16, which again demonstrates the effectiveness and superiority of multi-grained contrast and the AOSM module.
\end{itemize}

\begin{table}[]
\vspace{-0.2cm}
\caption{Retrieval performance comparison on MSVD.}
\vspace{-0.3cm}
\resizebox{1.00\columnwidth}{!}{
\begin{tabular}{l|ccc|ccc}
\hline
              & \multicolumn{3}{c|}{Text-to-Video }                                 & \multicolumn{3}{c}{Video-to-Text }                                 \\
              \hline
Model      & R@1↑          & R@5↑                 & MnR↓          & R@1↑          & R@5↑                   & MnR↓          \\ \hline
Multi Cues \cite{mithun2018learning}    & 20.3          & 47.8                   & -            & -             & -             & -                       \\
CE \cite{liu2019use}            & 19.8          & 49.0               & -            & -             & -             & -                      \\
SSB  \cite{patrick2020support}         & 28.4          & 60.0               & -            & -             & -             & -                      \\
NoiseE  \cite{amrani2020noise}      & 20.3          & 49.0                  & -            & -             & -                     & -            \\
CLIP-straight \cite{portillo2021straightforward} & 37.0          & 64.1               & -            & 59.9          & 85.2         & -            \\
Frozen \cite{bain2021frozen}       & 33.7          & 64.7                  & -            & -             & -             & -                  \\
TT-CE+  \cite{croitoru2021teachtext}       & 25.4          & 56.9              & -            & 27.1          & 55.3                 & -            \\
CLIP4Clip-MeanP (ViT-B/32) \cite{luo2021clip4clip} & 46.2          & 76.1          & 10.0         & 56.6          & 79.7          & 7.6          \\
% CLIP4Clip-seqLSTM (ViT-B/32) \cite{luo2021clip4clip}     & 46.2 & 75.3 & 10.2 & 52.5 & 74.0 & 14.7 \\
CLIP4Clip-seqTransf (ViT-B/32) \cite{luo2021clip4clip}   & 45.2 & 75.5 & 10.3 & 62.0 & 87.3 & 4.3  \\
% CLIP4Clip-tightTransf (ViT-B/32) \cite{luo2021clip4clip}  & 40.0 & 71.5 & 13.3 & 54.3 & 85.3 & 6.0  \\
CLIP4Clip-MeanP (ViT-B/16) \cite{luo2021clip4clip} & 47.3 & 77.7 & 9.1 & 62.9 & 87.2 & \textbf{4.2} \\ 
CLIP4Clip-seqTransf (ViT-B/16) \cite{luo2021clip4clip}   & 47.2 & 77.7 & 9.1 & 63.2 & 87.2 & \textbf{4.2}  \\
\hline
X-CLIP (ViT-B/32)    & 47.1          & 77.8          & 9.5          & 60.9          & 87.8          & 4.7          \\
X-CLIP (ViT-B/16)    & \textbf{50.4} & \textbf{80.6} & \textbf{8.4} & \textbf{66.8} & \textbf{90.4} & \textbf{4.2} \\ \hline
\end{tabular}
}
\vspace{-0.2cm}
\label{tab:msvd_performance}
\end{table}

\begin{table}[]
\vspace{-0.2cm}
\caption{Retrieval performance comparison on LSMDC.}
\vspace{-0.3cm}
\resizebox{1.00\columnwidth}{!}{
\begin{tabular}{l|ccc|ccc}
\hline
              & \multicolumn{3}{c|}{Text-to-Video}   & \multicolumn{3}{c}{Video-to-Text}   \\ \hline
Model         & R@1↑          & R@5↑          & MnR↓          & R@1↑          & R@5↑          & MnR↓          \\ \hline
CT-SAN  \cite{yu2017end}      & 5.1           & 16.3          & -             & -             & -             & -             \\
JSFusion \cite{yu2018joint}     & 9.1           & 21.2          & -             & 12.3          & 28.6          & -             \\
CE   \cite{liu2019use}         & 11.2          & 26.9          & 96.8          & -             & -             & -             \\
MMT \cite{gabeur2020multi}          & 12.9          & 29.9          & 75.0          & -             & -             & -             \\
NoiseE \cite{amrani2020noise}       & 6.4           & 19.8          & -             & -             & -             & -             \\
CLIP-straight \cite{portillo2021straightforward} & 11.3          & 22.7          & -             & 6.8           & 16.4          & -             \\
MDMMT \cite{dzabraev2021mdmmt}        & 18.8          & 38.5          & 58.0          & -             & -             & -             \\
Frozen \cite{bain2021frozen}       & 15.0          & 30.8          & -             & -             & -             & -             \\
HiT \cite{liu2021hit}          & 14.0          & 31.2          & -             & -             & -             & -             \\
TT-CE+  \cite{croitoru2021teachtext}      & 17.2          & 36.5          & -             & 17.5          & 36.0          & -             \\
CLIP4Clip-MeanP (ViT-B/32) \cite{luo2021clip4clip} & 20.7          & 38.9          & 65.3          & 20.6          & 39.4          & 56.7          \\
%  CLIP4Clip-seqLSTM (ViT-B/32) \cite{luo2021clip4clip}   & 21.6 & 41.8 & 58.0 & 20.9 & 40.7 & 53.9 \\
 CLIP4Clip-seqTransf (ViT-B/32)  \cite{luo2021clip4clip} & 22.6 & 41.0 & 61.0 & 20.8 & 39.0 & 54.2 \\
%  CLIP4Clip-tightTransf (ViT-B/32)  \cite{luo2021clip4clip}  & 18.9 & 37.8 & 61.6 & 17.4 & 36.7 & 65.3 \\
CLIP4Clip-MeanP (ViT-B/16) \cite{luo2021clip4clip} & 23.5          & 43.2          & 54.8          & 22.6          & 50.5          & 50.3          \\ 
CLIP4Clip-seqTransf (ViT-B/16) \cite{luo2021clip4clip} & 23.5          & 45.2          & 51.6          & 23.2          & 42.4          & 47.4          \\ 
\hline
X-CLIP (ViT-B/32)    & 23.3          & 43.0          & 56.0          & 22.5          & 42.2          & 50.7          \\
X-CLIP (ViT-B/16)    & \textbf{26.1} & \textbf{48.4} & \textbf{46.7} & \textbf{26.9} & \textbf{46.2} & \textbf{41.9} \\ \hline
\end{tabular}
}
\vspace{-0.2cm}
\label{tab:lsmdc_performance}
\end{table}

We also further validate the generalization of X-CLIP on MSVD, LSMDC, DiDeMo and ActivityNet in Tab. \ref{tab:msvd_performance} - \ref{tab:acnet_performance}. It is worth noting that, in all variants of CLIP4Clip, we only report the performance of \emph{CLIP4Clip-MeanP} and \emph{CLIP4Clip-seqTranf}, because they perform better than the other two variants in consideration of experience in the previous work \cite{luo2021clip4clip} and performance comparison in Tab. \ref{tab:mstvtt_performance}. By analyzing these tables, we can observe that X-CLIP also achieves significant improvement on these datasets for text-to-video and video-to-text retrieval tasks. Specifically, for the text-to-video retrieval task, X-CLIP outperforms the CLIP4Clip with ViT-B/16 on R@1 by +6.6\% (+3.1\%), +11.1\% (+2.6\%), +6.7\% (+3.0\%), +3.8\% (+1.7\%) relative (absolute) improvement on aforesaid four datasets respectively. For the video-to-text retrieval task, X-CLIP obtains +5.7\% (+3.6\%), +12.9\% (+3.0\%), +1.3\% (+0.6\%), +5.2\% (+2.3\%) relative (absolute) improvement on R@1. This demonstrates that our proposed X-CLIP can achieve consistent performance improvement on several video-text retrieval datasets. More experimental results are in the supplementary materials.

\begin{table}[]
\vspace{-0.2cm}
\caption{Retrieval performance comparison on DiDeMo.}
\vspace{-0.3cm}
\resizebox{1.00\columnwidth}{!}{
\begin{tabular}{l|ccc|ccc}
\hline
           & \multicolumn{3}{c|}{Text-to-Video}             & \multicolumn{3}{c}{Video-to-Text}            \\ \hline
Model      & R@1↑          & R@5↑          & MnR↓          & R@1↑          & R@5↑          & MnR↓         \\ \hline
S2VT \cite{venugopalan2014translating}      & 11.9          & 33.6          & -             & 13.2          & 33.6          & -            \\
FSE \cite{zhang2018cross}       & 13.9          & 36.0          & -             & 13.1          & 33.9          & -            \\
CE  \cite{liu2019use}        & 16.1          & 41.1          & 43.7          & 15.6          & 40.9          & 42.4         \\
ClipBERT \cite{lei2021less}  & 20.4          & 48.0          & -             & -             & -             & -            \\
Frozen \cite{bain2021frozen}    & 34.6          & 65.0          & -             & -             & -             & -            \\
TT-CE+  \cite{croitoru2021teachtext}   & 21.6          & 48.6          & -             & 21.1          & 47.3          & -            \\ 
CLIP4Clip-MeanP (ViT-B/32) \cite{luo2021clip4clip} & 43.4          & 70.2          & 17.5          & 42.5          & 70.6          & 11.6          \\
%  CLIP4Clip-seqLSTM (ViT-B/32) \cite{luo2021clip4clip}    & 43.4 & 69.9 & 17.5 & 42.4 & 69.2 & 11.8 \\
 CLIP4Clip-seqTransf (ViT-B/32) \cite{luo2021clip4clip}   & 42.8 & 68.5 & 18.9 & 41.4 & 68.2 & 12.4 \\
%  CLIP4Clip-tightTransf (ViT-B/32) \cite{luo2021clip4clip}  & 25.8 & 52.8 & 27.3 & 21.5 & 51.1 & 22.4 \\
CLIP4Clip-MeanP (ViT-B/16) \cite{luo2021clip4clip} & 44.8          & 75.1          & 13.0          & 47.2          & 74.0          & \textbf{10.5} \\ 
CLIP4Clip-seqTransf (ViT-B/16) \cite{luo2021clip4clip} & 44.8          & 73.4          & 13.5          & 44.7          & 74.0          & 10.6 \\ 
\hline
X-CLIP (ViT-B/32)    & 45.2          & 74.0          & 14.6          & 43.1          & 72.2          & 10.9          \\
X-CLIP (ViT-B/16)    & \textbf{47.8} & \textbf{79.3} & \textbf{12.6} & \textbf{47.8} & \textbf{76.8} & \textbf{10.5} \\ \hline
\end{tabular}
}
\vspace{-0.2cm}
\label{tab:didemo_performance}
\end{table}

\begin{table}[]
\vspace{-0.2cm}
\caption{Retrieval performance comparison on ActivityNet.}
\vspace{-0.3cm}
\resizebox{1.00\columnwidth}{!}{
\begin{tabular}{l|ccc|ccc}
 \hline
           & \multicolumn{3}{c|}{Text-to-Video}            & \multicolumn{3}{c}{Video-to-Text}            \\ \hline
Model      & R@1↑          & R@5↑          & MnR↓         & R@1↑          & R@5↑          & MnR↓         \\ \hline
FSE \cite{zhang2018cross}       & 18.2          & 44.8          & -            & 16.7          & 43.1          & -            \\
CE  \cite{liu2019use}        & 18.2          & 47.7          & 23.1         & 17.7          & 46.6          & 24.4         \\
HSE  \cite{zhang2018cross}      & 20.5          & 49.3          & -            & 18.7          & 48.1          & -            \\
MMT  \cite{gabeur2020multi}       & 28.7          & 61.4          & 16.0         & 28.9          & 61.1          & 17.1         \\
SSB   \cite{patrick2020support}       & 29.2          & 61.6          & -            & 28.7          & 60.8          & -            \\
HiT   \cite{liu2021hit}      & 29.6          & 60.7          & -            & -             & -             & -            \\
ClipBERT  \cite{lei2021less}  & 21.3          & 49.0          & -            & -             & -             & -            \\
TT-CE+   \cite{croitoru2021teachtext}   & 23.5          & 57.2          & -            & 23.0          & 56.1          & -            \\
CLIP4Clip-MeanP (ViT-B/32) \cite{luo2021clip4clip} & 40.5 & 72.4 & 7.4 & 42.5 & 74.1 & 6.6 \\
%  CLIP4Clip-seqLSTM (ViT-B/32) \cite{luo2021clip4clip}    & 40.1 & 72.2 & 7.3  & 42.6 & 73.4 & 6.7  \\
 CLIP4Clip-seqTransf (ViT-B/32) \cite{luo2021clip4clip}   & 40.5 & 72.4 & 7.5  & 41.4 & 73.7 & 6.7  \\
%  CLIP4Clip-tightTransf (ViT-B/32) \cite{luo2021clip4clip}  & 19.5 & 47.6 & 17.3 & 18.9 & 49.6 & 16.3 \\
CLIP4Clip-MeanP (ViT-B/16) \cite{luo2021clip4clip} & 44.0   &  73.9    &  7.0   &   44.1   &  74.0    &   6.5  \\ 
CLIP4Clip-seqTransf (ViT-B/16) \cite{luo2021clip4clip} & 44.5   &  75.2    &  \textbf{6.4}   &   44.1   &  75.2    &   \textbf{6.4} \\ 
\hline
X-CLIP (ViT-B/32)    & 44.3 & 74.1 & 7.9 & 43.9 & 73.9 & 7.6 \\
X-CLIP (ViT-B/16)    & \textbf{46.2} & \textbf{75.5} & 6.8 & \textbf{46.4} & \textbf{75.9} & \textbf{6.4}     \\   \hline
\end{tabular}
}
\vspace{-0.2cm}
\label{tab:acnet_performance}
\end{table}

\subsection{Ablation Study}

To fully examine the impact of different contrastive modules, we conduct an ablation study to compare different variants of X-CLIP. As shown in Tab. \ref{tab:module}, we gain two important observations:
\begin{itemize}[leftmargin=*]
    \item With the number of contrastive modules increasing, the retrieval performance tends to be higher. When X-CLIP is equipped with all contrastive modules, the best retrieval performance can be achieved. This may be because each contrastive module plays a different role in the retrieval task and different contrast modules can promote each other to achieve better retrieval results.
    
    \item  Our proposed cross-grained contrast can assist fine-grained contrast or coarse-grained contrast to achieve better performance in the retrieval task. Specifically, X-CLIP with the sentence-video contrast module (\emph{i.e.,}  Exp1) only achieves 43.0 R@1 in the text-to-video retrieval task. However, when X-CLIP is additionally equipped with cross-grained contrast modules (\emph{i.e.,}  Exp8 and Exp9), the performance gets obvious absolute improvements of 2.4\% and 1.0\% respectively. Similarly, when X-CLIP is only equipped with fine-grained and coarse-grained contrast modules (\emph{i.e.,}  Exp10), it achieves 44.8 R@1 in the text-to-video task. However, when it is additionally equipped with cross-grained contrast modules (\emph{i.e.,}  Exp13 and Exp14), 1.0\% and 0.7\% absolute improvement of R@1 can be achieved. Therefore, we conclude that the performance improvement of cross-grained contrast modules in the retrieval task does not conflict with that of coarse-grained and fine-grained contrast modules.

\end{itemize}

\begin{table*}[]
\caption{Retrieval performance with different contrastive granularity on the MSR-VTT dataset.}
\vspace{-0.3cm}
\begin{tabular}{c|c|c|c|c|cccc|cccc}
\hline
 & \multicolumn{4}{c|}{Contrastive Module}                & \multicolumn{4}{c|}{Text-to-Video}                             & \multicolumn{4}{c}{Video-to-Text}                            \\ \hline
ID&Sent-Video & Sent-Frame & Word-Video           & Word-Frame & R@1↑          & R@5↑          & R@10↑         & MnR↓          & R@1↑          & R@5↑          & R@10↑         & MnR↓         \\ \hline
Exp1& \checkmark       &           &                      &            & 43.0          & 70.7          & 81.6          & 16.3          & 43.0          & 70.2          & 81.2          & 11.5         \\
Exp2&         & \checkmark         &                      &            & 42.7          & 69.6          & 81.3          & 13.9          & 43.1          & 70.7          & 82.1          & 9.9          \\
Exp3&         &           & \checkmark                    &            & 42.8          & 69.9          & 80.1          & 17.0          & 43.2          & 70.1          & 80.5          & 13.8         \\
Exp4&         &           &                      & \checkmark          & 42.7          & 69.5          & 81.3          & 14.4          & 42.8          & 70.8          & 81.7          & 10.6         \\ \hline
Exp5&         & \checkmark         & \checkmark                    &            & 44.6          & 72.8          & 82.4          & 13.9          & 45.7          & 73.2          & 82.3          & \textbf{9.1}          \\
Exp6&         &           & \checkmark                    & \checkmark          & 45.6          & 72.0          & 82.0          & 13.6          & 44.8          & 72.5          & 81.7          & 9.6          \\
Exp7&         & \checkmark         &  & \checkmark          & 44.1          & 70.2          & 81.3          & 14.3          & 44.4          & 71.6          & 82.8          & 9.7          \\
Exp8& \checkmark       & \checkmark         &                      &            & 45.4          & 72.2          & 81.6          & 13.4          & 45.4          & 72.8          & 82.7          & 9.2          \\
Exp9& \checkmark       &           & \checkmark                    &            & 44.0          & 70.3          & 82.5          & 13.9          & 43.6          & 70.9          & 81.8          & 11.3         \\
Exp10& \checkmark       &           &                      & \checkmark          & 44.8          & 72.6          & 83.0          & 13.6          & 45.3          & 73.0          & 83.8          & 9.5          \\  \hline
Exp11&         & \checkmark         & \checkmark                    & \checkmark          & 45.7          & 72.7          & 82.5          & \textbf{13.2}          & 45.6          & 72.8          & 82.9          & 9.2          \\
Exp12& \checkmark       & \checkmark         & \checkmark                    &            & 45.7          & 72.7          & 82.5          & \textbf{13.2}          & 45.6          & 72.8          & 82.9          & 9.2          \\
Exp13& \checkmark       & \checkmark         &                      & \checkmark          & 45.8          & \textbf{73.2} & 82.7          & \textbf{13.2}          & 46.5          & 72.6          & 83.8 & 9.7 \\
Exp14& \checkmark       &           & \checkmark                    & \checkmark          & 45.5          & 72.8          & 82.9          & 13.5          & 46.4          & 72.5          & 83.7          & 9.6          \\ \hline
Exp15& \checkmark       & \checkmark         & \checkmark                    & \checkmark          & \textbf{46.1} & 73.0          & \textbf{83.1} & \textbf{13.2} & \textbf{46.8} & \textbf{73.3} & \textbf{84.0}          & \textbf{9.1}      \\    \hline
\end{tabular}
\label{tab:module}
\end{table*}

To justify the effectiveness of the proposed AOSM module, we compare our method with the conventional \emph{Mean-Max} and other variants (\emph{i.e.,} \emph{Max-Max}, \emph{Max-Mean} and \emph{Mean-Mean}). As shown in Tab. \ref{tab:attention}, we observe that the \emph{Mean-Mean} strategy performs worst. This may be because the \emph{Mean-Mean} strategy, which applies the same weight to all similarity scores during aggregating, can not eliminate the adverse effects of unnecessary frames and unimportant words on the retrieval results. The \emph{Max-Mean}, \emph{Mean-Max} and \emph{Max-Max} strategies perform better than the \emph{Mean-Mean} strategy. This can be attributed to that these strategies adopt the highest similarity during aggregation, so contrast scores between unnecessary frames and unimportant words will be filtered out. However, since these strategies adopt the top-1 similarity score, some important similarity scores will also be ignored. To address this issue, we propose the AOSM module, where all similarity scores will be applied with different weights during aggregation. From Tab. \ref{tab:attention}, we observe that compared with other strategies, our proposed attention mechanism achieves better performance.

\begin{table}[]
\vspace{-0.2cm}
\caption{Retrieval performance with different fusion methods for similarity matrices on the MSR-VTT dataset.}
\vspace{-0.3cm}
\resizebox{1.00\columnwidth}{!}{
\begin{tabular}{l|ccc|ccc}
\hline
\multicolumn{1}{c}{} & \multicolumn{3}{|c|}{Text-to-Video}             & \multicolumn{3}{c}{Video-to-Text}            \\ \hline
Method               & R@1↑          & R@5↑          & MnR↓          & R@1↑          & R@5↑          & MnR↓         \\ \hline
Max-Max              & 44.0          & 72.6          & 13.5          & 44.4          & 72.5          & 9.2          \\
Mean-Mean            & 43.2          & 71.2          & 14.8          & 42.5          & 70.2          & 11.4         \\
Mean-Max             & 44.4          & 71.1          & 14.9          & 44.2          & 71.7          & 10.2         \\
Max-Mean             & 44.9          & 71.3          & 13.5          & 43.8          & 71.8          & 9.4          \\
Attention            & \textbf{46.1} & \textbf{73.0} & \textbf{13.2} & \textbf{46.8} & \textbf{73.3} & \textbf{9.1} \\ \hline
\end{tabular}
}
\vspace{-0.2cm}
\label{tab:attention}
\end{table}

\begin{figure}
\centering 
  \includegraphics[width=0.9\columnwidth]{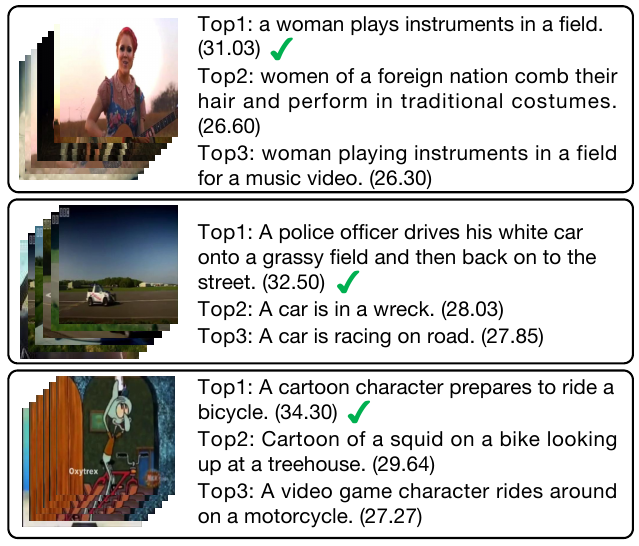}
  \vspace{-0.3cm}
  \caption{ Top-3 video-to-text retrieval results on MSR-VTT. The number in parentheses is the similarity score.
  }
  \vspace{-0.3cm}
  \label{fig:v2trank}
\end{figure}

\begin{table}[]
\caption{Ablation study of temporal encoder on the MSR-VTT dataset. \emph{TE} is short for temporal encoder.}
\vspace{-0.3cm}
\resizebox{1.00\columnwidth}{!}{
\begin{tabular}{l|c|ccc|ccc}
\hline
                          &                  & \multicolumn{3}{c}{Text-to-Video}             & \multicolumn{3}{c}{Video-to-Text}            \\ \hline
Base Model                & \emph{TE} & R@1↑          & R@5↑          & MnR↓          & R@1↑          & R@5↑          & MnR↓         \\ \hline
\multirow{2}{*}{ViT-B/32} &                  & 45.2          & 72.9          & 13.8          & 45.6          & 73.9          & 9.2          \\
                          & \checkmark                & \textbf{46.1} & \textbf{73.0} & \textbf{13.2} & \textbf{46.8} & \textbf{73.3} & \textbf{9.1} \\ \hline
\multirow{2}{*}{ViT-B/16} &   & 48.3               &      75.3         &       13.4        &         47.6      &        76.1                     &   9.0           \\
                          & \checkmark                & \textbf{49.3} & \textbf{75.8} & \textbf{12.2} & \textbf{48.9} & \textbf{76.8} & \textbf{8.1} \\\hline
\end{tabular}
}
\vspace{-0.2cm}
\label{tab:temporal}
\end{table}

To explore the impact of the temporal encoder module in X-CLIP, we also conduct an ablative study to compare the X-CLIP with and without the temporal encoder. As shown in Tab \ref{tab:temporal}, based on either ViT-B/32 or ViT/16, X-CLIP with temporal encoder consistently outperforms X-CLIP without temporal encoder. This may be because the temporal encoder is used to model the temporal relation of different frames in a video. Therefore, X-CLIP without temporal encoder can not understand and perceive the information that requires a combination of multiple frames, \emph{e.g.,} action. Based on the above analysis, we conclude that temporal modeling is also a key to improving the performance of retrieval tasks.

\subsection{Effect of Temperature Parameter}
\begin{table}[]
\vspace{-0.2cm}
\caption{Retrieval performance with different temprature parameters $\tau$ in Softmax on the MSR-VTT dataset.}
\vspace{-0.3cm}
\resizebox{1.00\columnwidth}{!}{
\begin{tabular}{l|ccc|ccc}
\hline
\multicolumn{1}{c}{} & \multicolumn{3}{|c|}{Text-to-Video}             & \multicolumn{3}{c}{Video-to-Text}            \\ \hline
$\tau$           & R@1↑          & R@5↑          & MnR↓          & R@1↑          & R@5↑          & MnR↓         \\ \hline
1          & 43.9          & 71.6          & 14.5          & 43.5          & 71.3          & 11.3          \\
0.1          & 45.2          & 72.2          & 14.0          & 45.3          & 73.1          & 9.3          \\
0.01          & \textbf{46.1} & \textbf{73.0} & \textbf{13.2} & \textbf{46.8} & \textbf{73.3} & \textbf{9.1} \\
0.001          & 45.6          & 72.2          & 13.7          & 43.6          & 72.5          & 9.4          \\ \hline
\end{tabular}
}
\vspace{-0.4cm}
\label{tab:tau}
\end{table}

To explore the effect of different $\tau$ in the AOSM module, we also designed a group of experiments by setting different temperature parameters $\tau$ in Softmax. From Tab. \ref{tab:tau}, we observe that the retrieval performance first improves before reaching the saturation point (\emph{i.e.,} $\tau = 0.01$), and then begins to decline slightly. The main reason may be that when $\tau$ is large, too many noisy similarity scores are considered. On the contrary, if the $\tau$ is small, some important similarity scores may be ignored. Besides, our proposed attention mechanism with different $\tau$ consistently performs better than the \emph{Mean-Mean} strategy, and the attention mechanism with the optimal $\tau$ outperforms other strategies in all evaluation protocols. This justifies that our proposed attention mechanism helps to strengthen the influence of important similarity scores and weaken the influence of noisy similarity scores, thus achieving better retrieval performance.

\subsection{Qualitative Analysis}

\begin{figure}
\centering 
\vspace{-0.2cm}
  \includegraphics[width=0.9\columnwidth]{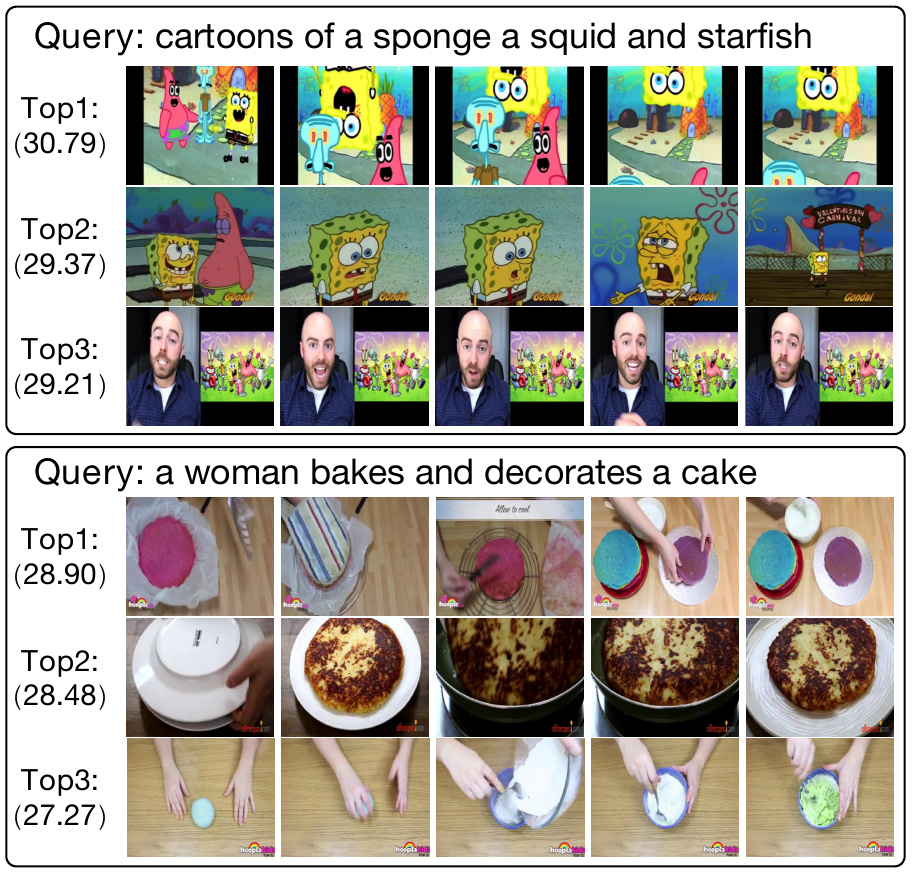}
  \vspace{-0.4cm}
  \caption{ Top-3 text-to-video retrieval results on MSR-VTT. The number in parentheses is the similarity score.
  }
  \vspace{-0.4cm}
  \label{fig:t2vrank}
\end{figure}

To qualitatively validate the effectiveness of our proposed X-CLIP, we show some typical video-to-text and text-to-video retrieval examples in Fig. \ref{fig:v2trank} and Fig. \ref{fig:t2vrank}, respectively. From these retrieval results, we find that X-CLIP could accurately understand the content of sentences and videos. Meanwhile, it is robust for X-CLIP to comprehend complex and similar sentences and videos, which is mainly attributed to the multi-grained contrast of our proposed model. To be specific, as shown in the first example in Fig.\ref{fig:v2trank}, although the top-3 retrieved sentences are similar, our proposed X-CLIP can still choose the correct sentence by understanding the details of sentences and videos. Similarly, as shown in the first example in Fig.\ref{fig:t2vrank}, all top-3 retrieved videos describe the same cartoon, while ``squid'' does not appear in the second and third videos. Due to the multi-grained contrast, X-CLIP performs well in visual and textual content understanding, so it can retrieve the correct video.

\section{Conclusion}
In this paper, we present X-CLIP, a novel end-to-end multi-grained contrastive model for video-text retrieval, which first encodes the sentences and videos into coarse-grained and fine-grained representations, and conducts fine-grained, coarse-grained, and cross-grained contrasts over these representations. The \emph{multi-grained contrast} and the \emph{AOSM} module of X-CLIP help to reduce the negative effects of unnecessary frames and unimportant words during retrieval. Significant performance gains on five popular video-text retrieval datasets demonstrate the effectiveness and superiority of our proposed model.

\begin{acks}
This work was supported by the National Science Fund for Distinguished Young Scholars (No.62025603), the National Natural Science Foundation of China (No. U21B2037, No. 62176222, No. 62176223, No. 62176226, No. 62072386, No. 62072387, No. 62072389, and No. 62002305), Guangdong Basic and Applied Basic Research Foundation (No.2019B1515120049), and the Natural Science Foundation of Fujian Province of China (No.2021J01002). This work was supported by Alibaba Group through Alibaba Research Intern Program.
\end{acks}

\bibliographystyle{ACM-Reference-Format}
\balance
\bibliography{sample-base}

\newpage

\section{Appendix}

\subsection{More Performance Comparison}

To verify the effectiveness of our method, we display the detailed comparison between our proposed X-CLIP and all variants of CLIP4Clip on different backbones (\emph{i.e.,} ViT-B/32 and ViT-B/16). As shown in  Tab. \ref{tab:msvd_performance2} - Tab. \ref{tab:acnet_performance2}, our proposed X-CLIP outperforms all variants of CLIP4Clip. Notably, X-CLIP with a weak backbone (\emph{i.e.,} ViT-B/32) even achieves comparable performance to CLIP4Clip with a strong backbone (\emph{i.e.,} ViT-B/16). This may be because our proposed cross-grained contrast is conducive to removing the noise information in the videos and sentences and capturing the important information. The outstanding performance again proves the importance and effectiveness of multi-grained contrast and the AOSM module.

\begin{figure}
\centering 
  \includegraphics[width=1\columnwidth]{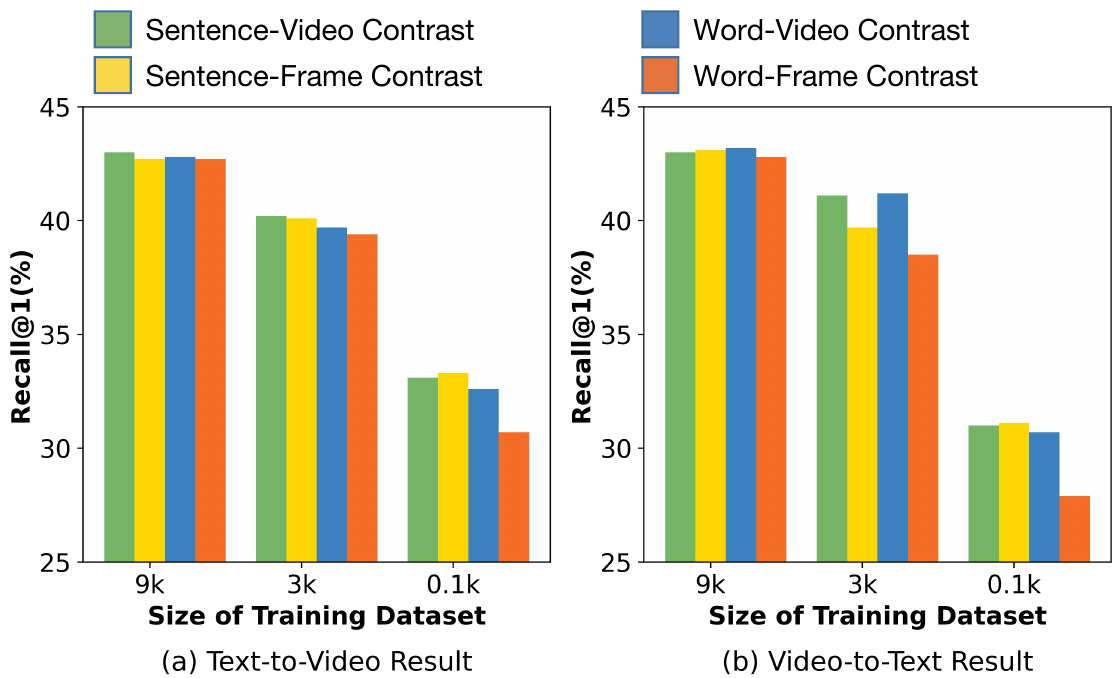}
  \caption{Retrieval performance of models with different contrastive modules in different sizes of the training set on the MSR-VTT dataset. }
  \label{fig:datasize}
\end{figure}

\begin{figure}
\centering 
  \includegraphics[width=1.0\columnwidth]{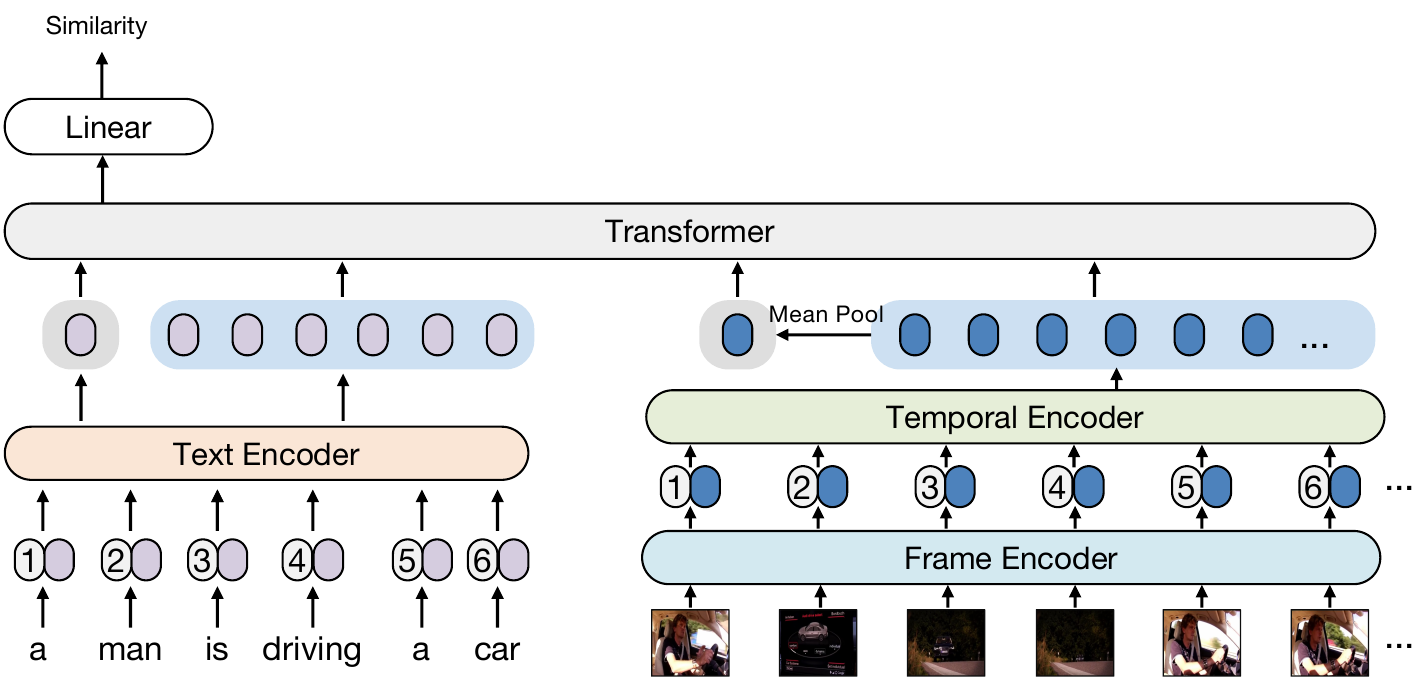}
  \caption{Overview of the new X-CLIP, which uses a Transformer to model multi-grained features rather than the AOSM module. }
  \label{fig:m_overview}
\end{figure}

\subsection{Effect of training dataset size on contrastive modules}

To gain deep insight into our four contrastive modules, we conduct the experiment to validate the X-CLIP with a single contrast module on the training datasets of different sizes. As illustrated in Fig. \ref{fig:datasize}, when the training data is sufficient (\emph{i.e.,} 9k), the video-to-text and text-to-video retrieval performance of four variants is similar. When the size of the training dataset is reduced to 3k, the performance differences of different variants begin to appear and the word-frame contrastive module performs worse than other modules. Furthermore, when the size of the training dataset is reduced to 0.1k, other contrastive modules perform better than the word-frame contrastive module by a significant margin. The main reason can be that compared with other modules, the word-frame contrastive module is more complex, so it is difficult to optimize this module on a small amount of training data.

\subsection{Effect of the AOSM module and Transformer modeling}     

To demonstrate the superiority and effectiveness of our proposed AOSM module, we also try to use a Transformer module to model the relationship of multi-grained features, which introduces more computation and parameters. The architecture of the new model is shown in Fig. \ref{fig:m_overview}. As shown in Tab. \ref{tab:Transf}, our proposed AOSM module performs better than the \emph{Transf} module. The performance gain can result from two aspects:

1) The new Transformer architecture introduces too many parameters, which makes it hard to be optimized with the limited amount of data. Our proposed AOSM is a well-designed module, where the importance of each frame and word is explicitly calculated. Thus, noise information in the video and sentence can be removed in X-CLIP. Besides, compared with Transformer, our proposed AOSM module contains fewer parameters, so it is easy to optimize the AOSM module.

2) The similarity scores of Transformer are obtained by a Linear layer, while the similarity scores of our proposed AOSM are obtained by the dot product. Notably, the dot product is the conventional approach for similarity calculation in the CLIP. However, Linear is a new approach, which does not carry any prior knowledge. Therefore, the prior knowledge of CLIP has little gain in Transformer, but our X-CLIP retains this prior knowledge well.

\begin{table*}[]
% \vspace{-0.2cm}
\caption{Retrieval performance comparison on MSVD.}
% \vspace{-0.3cm}
% \resizebox{1.00\columnwidth}{!}{
\begin{tabular}{l|ccc|ccc}
\hline
              & \multicolumn{3}{c|}{Text-to-Video }                                 & \multicolumn{3}{c}{Video-to-Text }                                 \\
              \hline
Model      & R@1↑          & R@5↑                 & MnR↓          & R@1↑          & R@5↑                   & MnR↓          \\ \hline
CLIP4Clip-MeanP (ViT-B/32)  & 46.2          & 76.1          & 10.0         & 56.6          & 79.7          & 7.6          \\
CLIP4Clip-seqLSTM (ViT-B/32)      & 46.2 & 75.3 & 10.2 & 52.5 & 74.0 & 14.7 \\
CLIP4Clip-seqTransf (ViT-B/32)    & 45.2 & 75.5 & 10.3 & 62.0 & 87.3 & 4.3  \\
CLIP4Clip-tightTransf (ViT-B/32)   & 40.0 & 71.5 & 13.3 & 54.3 & 85.3 & 6.0  \\
CLIP4Clip-MeanP (ViT-B/16)  & 47.3 & 77.7 & 9.1 & 62.9 & 87.2 & \textbf{4.2} \\ 
CLIP4Clip-seqLSTM (ViT-B/16)  & 48.4 & 78.0 & 9.1 & 61.2 & 87.6 & 5.0 \\ 
CLIP4Clip-seqTransf (ViT-B/16)    & 47.2 & 77.7 & 9.1 & 63.2 & 87.2 & \textbf{4.2}  \\
CLIP4Clip-tightTransf (ViT-B/16)    & 43.6 & 75.3 & 10.8 & 58.8 & 88.9 & 4.7  \\
\hline
X-CLIP (ViT-B/32)    & 47.1          & 77.8          & 9.5          & 60.9          & 87.8          & 4.7          \\
X-CLIP (ViT-B/16)    & \textbf{50.4} & \textbf{80.6} & \textbf{8.4} & \textbf{66.8} & \textbf{90.4} & \textbf{4.2} \\ \hline
\end{tabular}
% }
% \vspace{-0.2cm}
\label{tab:msvd_performance2}
\end{table*}

\begin{table*}[]
% \vspace{-0.2cm}
\caption{Retrieval performance comparison on LSMDC.}
% \vspace{-0.3cm}
% \resizebox{1.00\columnwidth}{!}{
\begin{tabular}{l|ccc|ccc}
\hline
              & \multicolumn{3}{c|}{Text-to-Video}   & \multicolumn{3}{c}{Video-to-Text}   \\ \hline
Model         & R@1↑          & R@5↑          & MnR↓          & R@1↑          & R@5↑          & MnR↓          \\ \hline
CLIP4Clip-MeanP (ViT-B/32)  & 20.7          & 38.9          & 65.3          & 20.6          & 39.4          & 56.7          \\
 CLIP4Clip-seqLSTM (ViT-B/32)    & 21.6 & 41.8 & 58.0 & 20.9 & 40.7 & 53.9 \\
 CLIP4Clip-seqTransf (ViT-B/32)   & 22.6 & 41.0 & 61.0 & 20.8 & 39.0 & 54.2 \\
 CLIP4Clip-tightTransf (ViT-B/32)    & 18.9 & 37.8 & 61.6 & 17.4 & 36.7 & 65.3 \\
CLIP4Clip-MeanP (ViT-B/16)  & 23.5          & 43.2          & 54.8          & 22.6          & 50.5          & 50.3          \\ 
CLIP4Clip-seqLSTM (ViT-B/16)  & 21.9          & 39.5          & 60.7          & 19.3          & 39.3          & 57.6          \\ 
CLIP4Clip-seqTransf (ViT-B/16)  & 23.5          & 45.2          & 51.6          & 23.2          & 42.4          & 47.4          \\ 
CLIP4Clip-tightTransf (ViT-B/16)  & 19.4          & 39.1          & 62.2          & 16.1          & 37.7          & 58.3          \\ 
\hline
X-CLIP (ViT-B/32)    & 23.3          & 43.0          & 56.0          & 22.5          & 42.2          & 50.7          \\
X-CLIP (ViT-B/16)    & \textbf{26.1} & \textbf{48.4} & \textbf{46.7} & \textbf{26.9} & \textbf{46.2} & \textbf{41.9} \\ \hline
\end{tabular}
% }
% \vspace{-0.2cm}
\label{tab:lsmdc_performance2}
\end{table*}

\begin{table*}[]
% \vspace{-0.2cm}
\caption{Retrieval performance comparison on DiDeMo.}
% \vspace{-0.3cm}
% \resizebox{1.00\columnwidth}{!}{
\begin{tabular}{l|ccc|ccc}
\hline
           & \multicolumn{3}{c|}{Text-to-Video}             & \multicolumn{3}{c}{Video-to-Text}            \\ \hline
Model      & R@1↑          & R@5↑          & MnR↓          & R@1↑          & R@5↑          & MnR↓         \\ \hline
CLIP4Clip-MeanP (ViT-B/32)  & 43.4          & 70.2          & 17.5          & 42.5          & 70.6          & 11.6          \\
 CLIP4Clip-seqLSTM (ViT-B/32)     & 43.4 & 69.9 & 17.5 & 42.4 & 69.2 & 11.8 \\
 CLIP4Clip-seqTransf (ViT-B/32)    & 42.8 & 68.5 & 18.9 & 41.4 & 68.2 & 12.4 \\
 CLIP4Clip-tightTransf (ViT-B/32)   & 25.8 & 52.8 & 27.3 & 21.5 & 51.1 & 22.4 \\
CLIP4Clip-MeanP (ViT-B/16)  & 44.8          & 75.1          & 13.0          & 47.2          & 74.0          & \textbf{10.5} \\ 
CLIP4Clip-seqLSTM (ViT-B/16)  & 44.7 & 72.2 & 15.5 & 43.9 & 72.5 & 11.8  \\ 
CLIP4Clip-seqTransf (ViT-B/16)  & 44.8          & 73.4          & 13.5          & 44.7          & 74.0          & 10.6 \\ 
CLIP4Clip-tightTransf (ViT-B/16)  & 34.8 & 65.8 & 20.5 & 36.5 & 65.5 & 13.9 \\ 
\hline
X-CLIP (ViT-B/32)    & 45.2          & 74.0          & 14.6          & 43.1          & 72.2          & 10.9          \\
X-CLIP (ViT-B/16)    & \textbf{47.8} & \textbf{79.3} & \textbf{12.6} & \textbf{47.8} & \textbf{76.8} & \textbf{10.5} \\ \hline
\end{tabular}
% }
% \vspace{-0.2cm}
\label{tab:didemo_performance2}
\end{table*}

\begin{table*}[]
% \vspace{-0.2cm}
\caption{Retrieval performance comparison on ActivityNet.}
% \vspace{-0.3cm}
% \resizebox{1.00\columnwidth}{!}{
\begin{tabular}{l|ccc|ccc}
 \hline
           & \multicolumn{3}{c|}{Text-to-Video}            & \multicolumn{3}{c}{Video-to-Text}            \\ \hline
Model      & R@1↑          & R@5↑          & MnR↓         & R@1↑          & R@5↑          & MnR↓         \\ \hline
CLIP4Clip-MeanP (ViT-B/32)  & 40.5 & 72.4 & 7.4 & 42.5 & 74.1 & 6.6 \\
 CLIP4Clip-seqLSTM (ViT-B/32)     & 40.1 & 72.2 & 7.3  & 42.6 & 73.4 & 6.7  \\
 CLIP4Clip-seqTransf (ViT-B/32)    & 40.5 & 72.4 & 7.5  & 41.4 & 73.7 & 6.7  \\
 CLIP4Clip-tightTransf (ViT-B/32)   & 19.5 & 47.6 & 17.3 & 18.9 & 49.6 & 16.3 \\
CLIP4Clip-MeanP (ViT-B/16)  & 44.0   &  73.9    &  7.0   &   44.1   &  74.0    &   6.5  \\ 
CLIP4Clip-seqLSTM (ViT-B/16)  & 44.4 & 74.9 & 6.4 & 44.7 & 75.1 & 6.3  \\ 
CLIP4Clip-seqTransf (ViT-B/16)  & 44.5   &  75.2    &  \textbf{6.4}   &   44.1   &  75.2    &   \textbf{6.4} \\ 
CLIP4Clip-tightTransf (ViT-B/16)  & 30.8 & 64.3 & 9.8 & 29.6 & 62.3 & 9.9 \\ 
\hline
X-CLIP (ViT-B/32)    & 44.3 & 74.1 & 7.9 & 43.9 & 73.9 & 7.6 \\
X-CLIP (ViT-B/16)    & \textbf{46.2} & \textbf{75.5} & 6.8 & \textbf{46.4} & \textbf{75.9} & \textbf{6.4}     \\   \hline
\end{tabular}
% }
% \vspace{-0.2cm}
\label{tab:acnet_performance2}
\end{table*}

\begin{table*}[]
\caption{Retrieval performance comparison between Transformer modeling and the AOSM module. \emph{Transf} means using a 3-layer Transformer to model multi-grained features.}
% \resizebox{1.00\columnwidth}{!}{
\begin{tabular}{l|ccc|ccc} 
\hline
                    & \multicolumn{3}{c}{Text-to-Video Retrieval} & \multicolumn{3}{c}{Video-to-Text Retrieval}               \\ \hline
Model               & R@1↑          & R@5↑                 & MnR↓          & R@1↑          & R@5↑                   & MnR↓    \\ \hline
Trans(ViT-B/32)     & 38.9         & 68.8        & 13.9           & 39.4                & 69.4                & 11.6          \\
X-CLIP(ViT-B/32) & 46.1         & 73.0        & 13.2           & 46.8                & 73.3                & 9.1           \\ \hline
Trans(ViT-B/16)     & 40.1         & 70.5        & 14.0           & 41.4                & 71.5                & 10.8          \\
X-CLIP(ViT-B/16) & 49.3         & 75.8        & 12.2           & 48.9                & 76.8                & 8.1   \\ \hline       
\end{tabular}
% }
\label{tab:Transf}
\end{table*}

\end{document}